\documentclass[conference]{IEEEtran}
\IEEEoverridecommandlockouts
\usepackage{cite}
\usepackage{amsmath,amssymb,amsfonts}
\usepackage{algorithmic}
\usepackage{graphicx}
\usepackage{textcomp}
\usepackage{xcolor}
\usepackage{hyperref}

\usepackage{color}
\usepackage{array}
\usepackage{booktabs}
\usepackage{colortbl}

\usepackage{subfigure}
\usepackage{multirow}

\newcommand{\R}{\mathbb{R}}
\DeclareMathOperator*{\softmax}{softmax}
\DeclareMathOperator*{\advloss}{AdvLoss}

\usepackage{bbm}    
\usepackage{mathtools}
\usepackage{makecell}
\usepackage{hhline}

\def\BibTeX{{\rm B\kern-.05em{\sc i\kern-.025em b}\kern-.08em
    T\kern-.1667em\lower.7ex\hbox{E}\kern-.125emX}}
\begin{document}

\bibliographystyle{plain}

\title{Adversarial Mixture Of Experts with Category Hierarchy Soft Constraint}

\author{\IEEEauthorblockN{Zhuojian Xiao, Yunjiang Jiang, Guoyu Tang, Lin Liu, Sulong Xu, Yun Xiao, Weipeng Yan}
\IEEEauthorblockA{
\{xiaozhuojian5,yunjiang.jiang,tangguoyu,liulin1,xusulong,xiaoyun1,Paul.yan\}@jd.com
}
}

\maketitle

\begin{abstract}
Product search is the most common way for people to satisfy their shopping needs on e-commerce websites. Products are typically annotated with one of several broad categorical tags, such as “Clothing” or “Electronics”, as well as finer-grained categories like “Refrigerator” or “TV”, both under “Electronics”. These tags are used to construct a hierarchy of query categories. Distributions of features such as price and brand popularity vary wildly across query categories. In addition, feature importance for the purpose of CTR/CVR predictions differs from one category to another. In this work, we leverage the Mixture of Expert (MoE) framework to learn a ranking model that specializes for each query category. In particular, our gate network relies solely on the category ids extracted from the user query. 

While classical MoE’s pick expert towers spontaneously for each input example, we explore two techniques to establish more explicit and transparent connections between the experts and query categories. To help differentiate experts on their domain specialties, we introduce a form of adversarial regularization among the expert outputs, forcing them to disagree with one another. As a result, they tend to approach each prediction problem from different angles, rather than copying one another. This is validated by a much stronger clustering effect of the gate output vectors under different categories. In addition, soft gating constraints based on the categorical hierarchy are imposed to help similar products choose similar gate values. and make them more likely to share similar experts. This allows aggregation of training data among smaller sibling categories to overcome data scarcity. 

Experiments on a learning-to-rank dataset collected from the JD e-commerce search log demonstrate that MoE with these improvements consistently outperforms competing models, in terms of offline metrics and online AB tests.
\end{abstract}


\section{Introduction}
Increasingly people are turning to e-commerce to satisfy their shopping needs. From the early days of selling books and durable goods, today e-commerce platforms offer a wide range of products, including perishables and services. This poses fresh challenges in search ranking as the user queries invariably become more diverse and colloquial, similar to how users would interact with a store cashier.

One key input in e-commerce search ranking is the product category tagging. Often the shop owners are required to label their products with these categories, to facilitate search indexing. From these product categories one can construct a notion of query categories, usually by aggregating the most frequently occurring product categories correctly retrieved under the query. Most e-commerce ranking systems today do not have the engineering resource to deploy dedicated models for each query category, even the major ones. But as a human cataloguer, a natural strategy is to first identify the most likely category the query belongs to, then retrieve items within the category. 
Features in various categories may have different importance for product ranking. Intuitively, it is expected that separate ranking strategies on different categories should be able to improve overall product search relevance, as judged by user purchase feedback.

In order to put this intuition into practice without incurring unwieldy engineering cost, modeling ideas such as Mixture of Experts quickly come to mind. The latter excels at delegating a single task into a bidding and polling system of multiple expert predictors. The actual mechanism of MoE models however differs from this intuition: the model actually learns the experts spontaneously, without meaningful connection to natural covariates like the product category. While the model quality may be improved, the model is still opaque and monolithic, difficult to understand from the business dimension. 

Here we propose a set of techniques based on MoE to take advantage of natural business categories, such as electronics or books, which ultimately improves ranking quality on individual categories as well as making the expert specialties more distinctive and transparent.

\begin{figure}
    \centering
    \includegraphics[width=\linewidth]{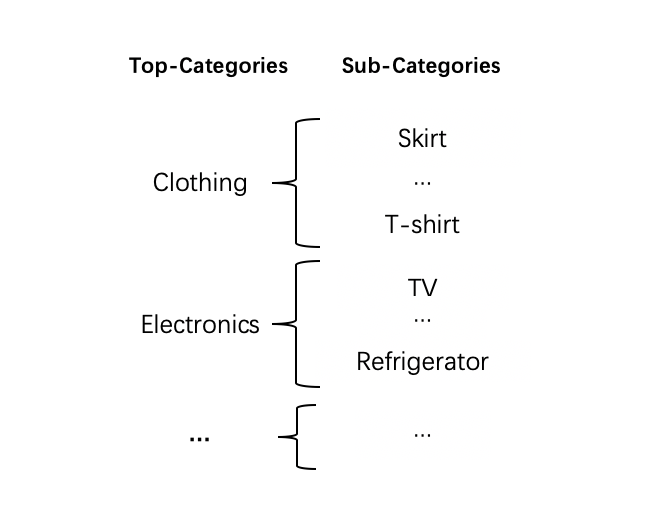}
    \caption{Hierarchical Categories}
    \label{fig:hierarchy-category}
\end{figure}

We summarize our contributions as follows:

\begin{itemize}
    \item \textbf{Hierarchical Soft Constraint:} We introduce a novel soft constraint based on hierarchical categories (Figure~\ref{fig:hierarchy-category}) in the e-commerce product search scenario to help similar categories learn from one another. By sharing network weights more strategically among similar categories, smaller sibling categories can combine their training data to mitigate data size skew.
    \item \textbf{Adversarial Mixture of Experts:} We propose an adversarial regularization technique in MoE model to encourage that experts of different problem domains disagree with one another, thereby improving diversity of viewpoints in the final ensemble.
    \item \textbf{Benefits in Real-world Datasets:} To our best knowledge, this work represents the first study of deep MoE models on learning-to-rank datasets. An early work based on classical modeling techniques can be found in \cite{chakraborty2007learning}. Applications in content recommendation domains such as \cite{mmoe} do not involve the user query, which is a key input feature in our models. Experiments show that our improved MoE model outperforms competing methods, especially on smaller categories that have traditionally suffered from insufficient training data.
\end{itemize}

\section{Related Works}


\subsection{Deep Search and Recommendation Algorithms}
Since the beginning of the neural net revolution, deep neural nets have been successfully applied in industrial ranking problems, with the notable pioneer of DSSM \cite{huang2013learning} that embeds query and documents as separate input features in a multi-layer perception.

Subsequent improvements of DSSM include Deep \& Cross \cite{deep-cross}, Wide \& Deep \cite{wide-deep}, DeepFM \cite{deepfm}, etc. A fully-connected network approach is presented in \cite{li2019semantic}. Reinforcement learning in conjunction with DNN in the e-commerce search setting has been extensively studied in \cite{hu2018reinforcement}. Ensembling of neural net models with other traditional techniques has been explored in \cite{wu2017ensemble}. Mixture of Experts can also be viewed as a form of end-to-end ensembling, with potentially unbounded number of ensemble components at constant serving cost.


The invention of attention-based transformer model \cite{DBLP:journals/corr/VaswaniSPUJGKP17} ushered in a new era of natural language based information retrieval system, most notably BERT \cite{devlin2018bert} and its variant successors. These have been applied in large scale industrial text search system \cite{nayak_2019} to dramatically improve textual relevance. E-commerce search however has an additional emphasis on user conversion rate, thus the use of non-text features is essential, and typically requires training from scratch with custom dataset.

An orthogonal angle to our notion of query categories is presented in \cite{sondhi2018taxonomy}. The authors classify queries into 5 generic categories based on user intent, while we try to match the experts with existing product categories.

In the content recommendation domain, we cite some recent exciting advancements exploiting users' historical behavior sequence, most notably DIN \cite{zhou2018deep}, DIEN \cite{naumov2019deep}, and MIMN \cite{pi2019practice}. Here the entire user history and profile take on a similar role as the query. We do not focus particularly on user history in this work. This makes our method more generally applicable. Our only requirement is that each query be assigned a category id, preferably with hierarchical structure.

\subsection{Mixture of Experts} \label{moe_review}
 The idea of Mixture of Experts (MoE) was first introduced in \cite{jacobs1991adaptive}, and has been used in a considerable number of Machine Learning (ML) problems, such as pattern recognition and classification; see \cite{twentyyearsmoe} for an early survey. The idea proposed in \cite{jacobs1991adaptive} is quite intuitive: a separate gate network (usually a single layer neural net) softly divides a complex problem into simpler sub-spaces, to be assigned to one or more expert networks. Different ML models have been used as expert network and gate network in MoE, such as neural networks model \cite{jordan1994hierarchical}, SVM \cite{LIMA20072049}, \cite{CAOsvm}, and decision trees \cite{arnaud2019tree}. Above work shares many similarities, namely that the complete model consists of many expert sub-models and one or a few gate models that control the activation of the experts. \cite{deepmoe} introduces a deep mixture of experts by stacking multiple sets of gating and experts as multiple layer MoE. Model capacity of deep mixture of experts increases polynomially, through different combinations of experts in each layer. 

 A major milestone in language modeling applications appears in \cite{shazeer2017outrageously}, which introduces a novel gate network, called Noisy Top-K Gating, to accommodate both large model capacity and efficient computation. Since the gate only activates a few experts ($K \ll N$ the total number of experts) for each example, the model can increase capacity by increasing a large number of experts. Furthermore, their model imposes a load balancing regularization constraint so that each expert sees roughly the same number of training examples per mini-batch, and gets similar share of training opportunities. In this work, we extend the Noisy Top-K Gating as well as the load-balancing idea with a hierarchical soft constraint to help similar examples share the same experts. In addition, we also encourage experts to disagree among themselves, so that various experts specialize in different problem sub-spaces.
 
 One major variant of \cite{shazeer2017outrageously} is the MMoE (Multi-gate Mixture-of-Experts) model \cite{mmoe}, which combines multi-task learning with MoE, by allocating multiple independent gate networks, one for each task. This was successfully deployed in the Youtube recommendation system \cite{mmoe2019}. One of our baseline models attempts to replicate this by treating different groups of major product categories as different tasks, within a mini-batch.

Finally, we mention two recent applications of MoE. 
\cite{wang2018deep} embeds hundreds of shallow MoE networks, one for each layer of a convolutional neural net, as a way to reduce its computational cost while maintaining or improving generalization error. This is however mostly applicable in the computer vision domain.

In the video captioning domain, \cite{wang2019learning} builds an MoE model on top of an underlying recurrent neural net with attention mechanism, and achieves impressive wins in generating accurate captions for previously unseen video activities.

\subsection{Ensemble Regularization}
The idea of encouraging ensemble disagreement has been explored in \cite{li-etal-2018-multi-head}, in the context of multi-head attention. The latter applies disagreement regularization in three different parts of the transformer unit. Our adversarial regularization is more straightforward and applies only to the output of the experts. Furthermore, instead of applying the regularization to all pairs of experts, only those with the biggest gating score contrast are regularized. 

Hierarchical regularization has been studied in the context of multi-domain sentiment analysis \cite{xu2020hierarchical}. An earlier work \cite{zweig2013hierarchical} applies regularization to discover the optimal hierarchical relation among a multitude of classification tasks. While these works exploit hierarchical structure in the problem domain, to the best of our knowledge, regularization based on hierarchical structure in categorical feature spaces has not been systematically studied.

\section{Model Description}
 Formally, let $(X, y)$ to be an individual example in the training data set, where $X$ consists of product features, query/user features, and joint 2-sided features. Those features can be divided into two types: numeric features and sparse features; see Table~\ref{tab:feature_examples} for some examples.
 In practice, we employ an embedding procedure to transfer sparse features into dense vectors. Finally, we concatenate the embedding vectors and normalized numeric features as one input vector to the ranking model.
\begin{equation}
    \label{eq:input}
        X = [x_{e,1}^T, \dots, x_{e,k}^T, x_{d,1}, \dots, x_{d,m}] \in \R^{1 \times n}
\end{equation}
Here, $x_{e,i} \in \R^q$ is the embedding vector for the $i$-th sparse feature, and $x_{d,j}$ is the normalized value for $j$-th numeric features. $n = kq + m$ is the input layer width, where $q$ is the embedding dimension for each sparse feature. $X \in \R^{1 \times n}$ is thus the input vector fed to the rest of the model. $y \in \{0, 1\}$ indicates whether the user has purchased the product. Our goal is to learn a ranking model to evaluate the purchase probability of a product for a given query.

\begin{table}
    \centering
    \caption{Feature Examples}
    \begin{tabular}{|c|c|c|}
        \hline
        ~ & Numeric & Sparse \\
        \hline
        \multirow{3}*{Product Side} & Click-Through Rate & Ware ID \\
        ~ & Conversion Rate & Sub-Category ID \\
        ~ & Price & Top-Category ID \\
        ~ & Sale Volume & Product Title \\
        \hline
        \multirow{3}*{User/Query Side} & User Age & Query Text  \\
        ~ & User Purchase Power & User ID \\
        ~ & User Activeness & Click History IDs \\
        \hline
        \multirow{2}*{Joint 2-Sided} & Model Relevance Score & - \\
        ~ & User/Sub-Category CTR & - \\
        \hline
    \end{tabular}
    \label{tab:feature_examples}
\end{table}

\begin{figure*}
    \centering
    \includegraphics[width=0.9\linewidth]{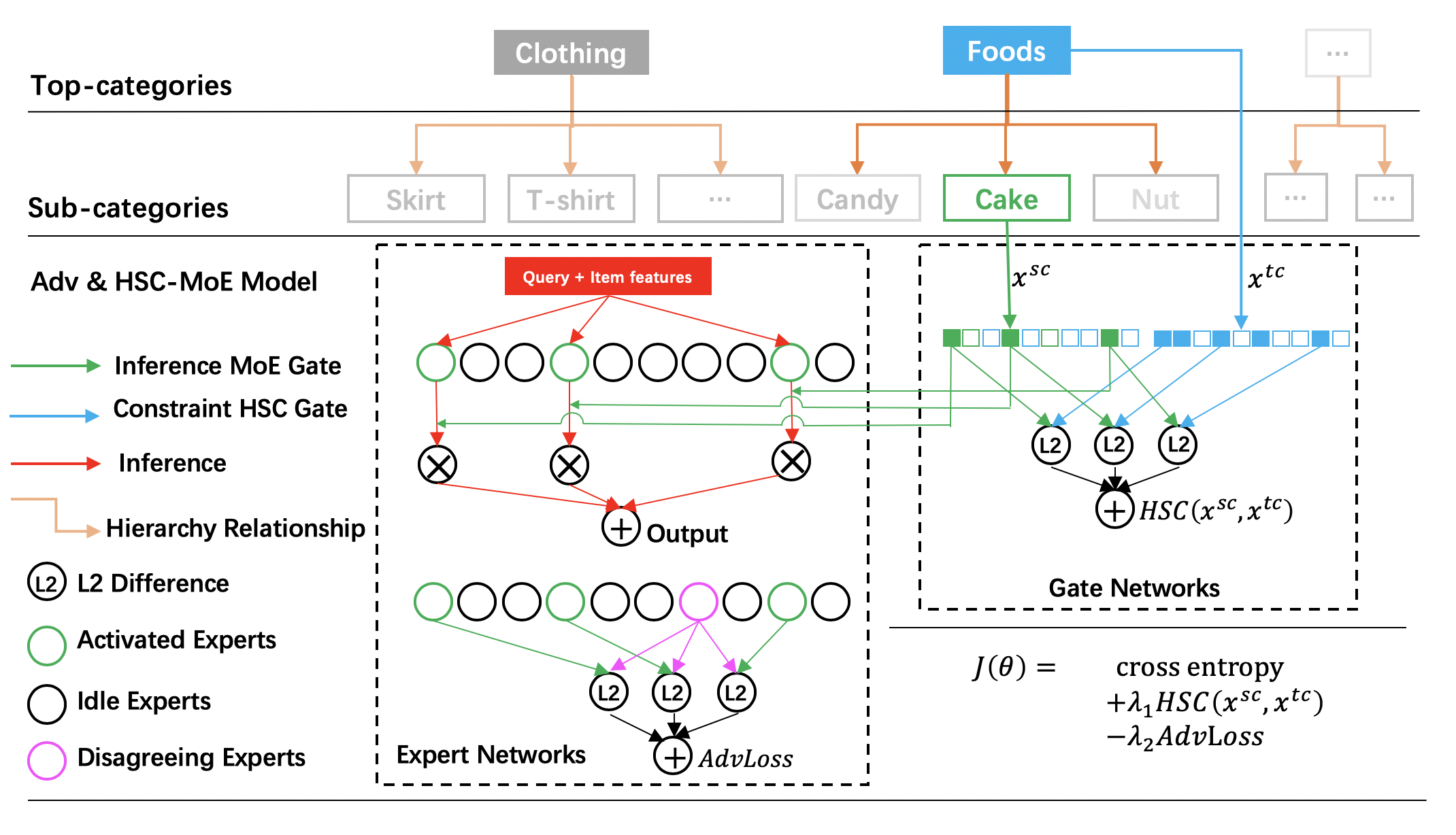}
    \caption{The architecture of Adv \& HSC-MoE.}
    \label{fig:adv-hsc-moe}
\end{figure*}

\subsection{Query Level product categorical Ids}

While most products have well-defined category ids, it is often helpful in a search engine to assign categories to queries as well, for instance to quickly filter irrelevant products during the retrieval phase. Since such features are crucial as our gating input, we describe their calculation in some detail.

First, we sample about 100k unique queries from the search log. These queries are then annotated with appropriate categories by multiple human raters, and cross validated to ensure consistency. A bidirectional GRU model is then trained with a softmax output layer to predict the most likely product category a given input query belongs to. The set of product categories is also pre-determined and updated very infrequently.

Once the model predicts the sub-categories for a given query, the top-categories are determined automatically via the category hierarchy.

\subsection{Basic Mixture of Experts Model}
We follow closely the Mixture of Experts model proposed in \cite{shazeer2017outrageously}, which appears to be its first successful application in a large scale neural network setting. 
\begin{figure}
    \centering
    \includegraphics[width=\linewidth]{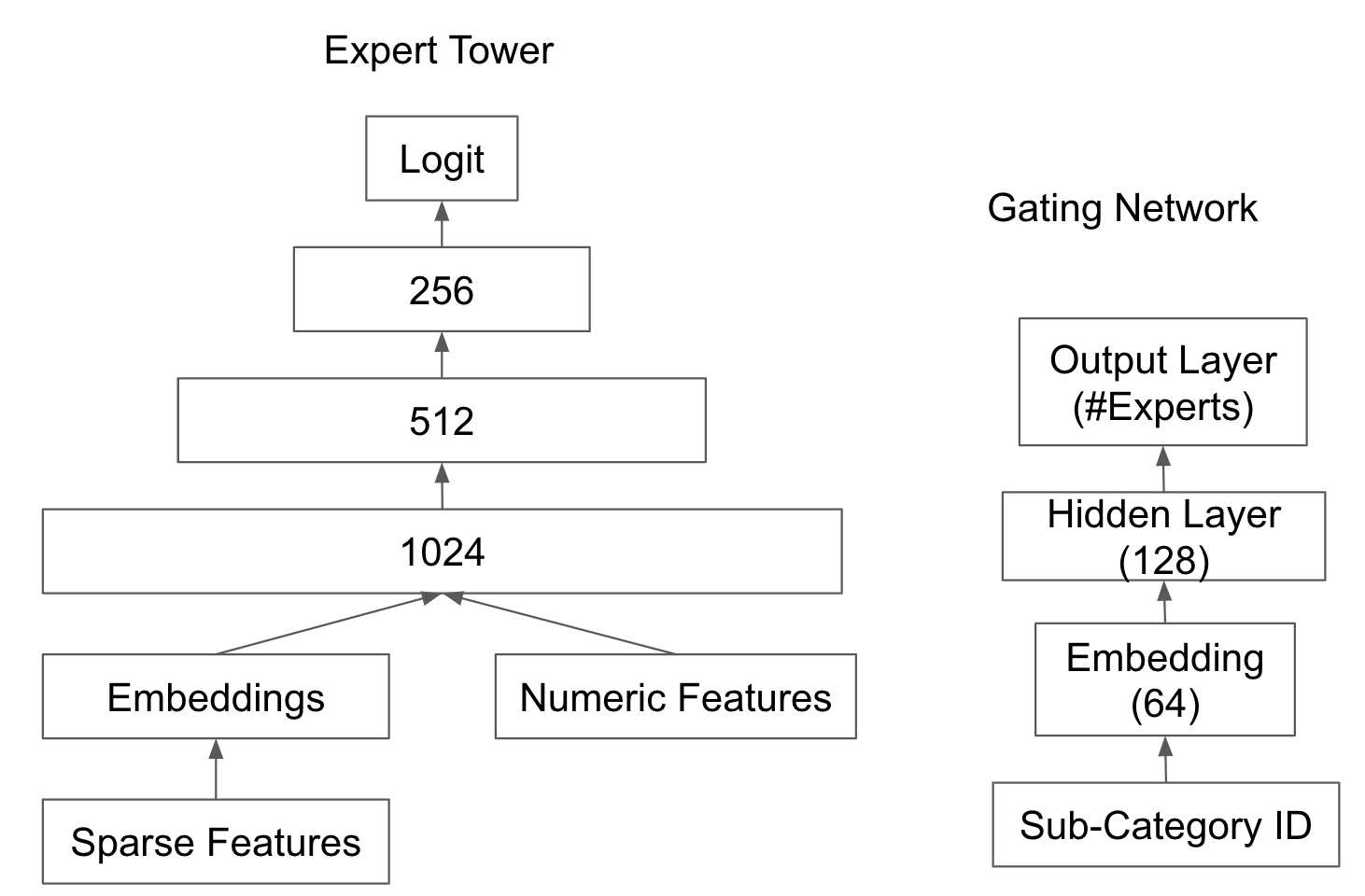}
    \centering
    \caption{Model sizes for each expert tower and the gating network.}
    \label{fig:model_sizes}
\end{figure}
The latter consists of a set of $N$ expert towers, $E_1, E_2, \dots, E_N$, with identical network structure (3 ReLU layers of sizes [1024, 512, 256] each) and different parameters due to random initialization, and a top K gate network. The output layer combines predictions from a subset of the experts selected by the gate network, weighted by their corresponding gate values. In this paper, we apply a Multi-Layer Perceptron (MLP) as the basic structure of an expert, due to its serving efficiency and ability to handle input data without any sequence structure. See Figure~\ref{fig:model_sizes} for detailed parameter settings. 

 Let $\Theta_g$ denote the gate network and $\Theta_{e_i}$, $1 \le i \le N$, the set of expert towers respectively. The MoE model predicts according to the following recipe:
 
 \begin{align}
 G &= \Theta_g(x^{sc}) \in \R^{1 \times N} \\
\hat{y} &= \sum_{1 \le i \le N: G_i \geq \hat{G}_K } G_i \Theta_{e_i}(X) \in \R
 \end{align}
 Here $x^{sc}$ stands for the input feature (in dense form) to the gate network, and $\hat{G}_K$ stands for the $K$-th largest gate value among $G_1, \ldots, G_N$, that is, $\mid\{i: G_i \geq \hat{G}_K\}\mid = K$. The final prediction logit is simply a weighted sum of the expert towers. Only the towers whose gate values are among the top $K$ will be computed during training and serving.



\subsection{Hierarchical Soft Constraint Gate}

The Hierarchical Soft Constraint Gate (HSC Gate) is an additional network, identical in structure to the base MoE gate network (MoE Gate), as proposed in \cite{shazeer2017outrageously}, namely a Noisy Top-K Gating. HSC Gate takes the top-categorical (TC) Ids as input, which by design is determined by the sub-category (SC) Ids, and therefore omitted from the input of MoE Gate, since the latter always takes sub-category Ids as one of the inputs.

As shown in Figure~\ref{fig:adv-hsc-moe}, TC and SC have a hierarchical relationship in a tree-based category system, where TC is a parent node and SC are children nodes.
To further emphasize the functions of gates, we call them \textbf{inference MoE gate} (green marked) $G^I$ and \textbf{constraint HSC gate} (blue marked) $G^C$ in the legend respectively.

\subsubsection{Inference MoE Gate}
 In our model, we feed the embedding vector of Sub-Category Ids $x^{sc}$ to the inference gate. Inference gate is designed as a parameterized indicator, whose output represents the weights of experts. We define the inference gate function as follows

\begin{equation}
    \label{eq:inference_gate}
    G^{I}(x^{sc}) = x^{sc} \cdot W^I \in \R^{1 \times N}
\end{equation}

where $W^I$ is a $q \times N$ dimensional weight matrix, $q$, $N$ being the embedding dimension and number of experts respectively. $G^I_i$ stands for the weight of the $i$-th expert. $x^{sc} \in X$ is SC embedding vector, a part of all input vector defined in \eqref{eq:input}. $W^I \in \R^{d_{emb} \times N}$ is a trainable matrix. 

To save computation, we only keep the top $K$ values in $G^I$ and set the rest to $-\infty$. Then we apply the softmax function to get the probability distributions from top $K$ $G^I_i$'s as follows.

\begin{align}
\label{eq:top_k}
&\tilde{G}_K^I(x^{sc}) = 
\left\{
    \begin{array}{ll}
    \d G^I_i(x^{sc}) & \text{if $G^I_i(x^{sc}) > \hat{G}_K(x^{sc})$}, \\
    \d -\infty & \text{otherwise}.
    \end{array}
\right. \\
    \label{eq:gate_prob}
    &P(x^{sc}, K) = \softmax(\tilde{G}_K^I(x^{sc})) \in \R^{1 \times N}
\end{align}
 As a result, only the probabilities of the top $K$ values remain greater than $0$. A noise term is added to the original output of $G^I$ to ensure differentiability of the top K operation, as detailed in \cite{shazeer2017outrageously}.
 
Like other MoE-based models, the output of our model can be written as follows \eqref{eq:y}. Since $P(x^{sc}, K)$ has $K$ values greater 0, we only activate these corresponding experts to save computation. 
The computational complexity of the model depends on the network of single expert and the value of $K$. 

\begin{equation}
    \label{eq:y}
    \hat{y} = \sum_{i=1}^N P_i(x^{sc}, K) E_i(X)
\end{equation}

\subsubsection{Constraint HSC Gate}
In our model, the constraint gate and inference gate have the same structure. We denote the constraint gate by $G^C$. In contrast to the inference gate $G^I$, however, the input feature of $G^C$, denoted $x^{tc}$, is the embedding vector of TC, which has a hierarchical relationship with the $x^{sc}$. As shown in Figure~\ref{fig:adv-hsc-moe} we define the Hierarchical Soft Constraint (HSC) between inference gate and constraint gate as follows:

\begin{align}
    \label{eq:gate_c_1}
    p^I(x^{sc}) &= \softmax(G^I(x^{sc})) \\
    \label{eq:gate_c_2}
    p^C(x^{tc}) &= \softmax(G^C(x^{tc})) \\
    \label{eq:gate_c_3}
    HSC(x^{sc}, x^{tc}) &= \sum_{i \in U_{top\_K}} (p^I(x^{sc})_i - p^C(x^{tc})_i)^2
\end{align}

Where $U_{top\_K}$ is the index set corresponding to top $K$ values in $G^I(x^{sc})$. $p^I \in \R^{1 \times N}$ and $p^C \in \R^{1 \times N}$ are probability distributions of inference gate and constraint gate.

By design, products from different sub-categories under the same top-categories are a lot more similar than products from completely different top-categories.
Therefore, it is intuitively helpful to share expert towers among sibling sub-categories. However we do not know a priori which experts to assign to each sub-category. Indeed the philosophy of MoE is to let the model figure it out by itself. On the other hand, we do not care about the exact experts assigned to each sub-category. The HSC gate thus seeks to preserve the hierarchy relationship between SC and TC, encouraging queries from sibling sub-categories to choose similar experts.  

HSC will be a part of loss function in our model to help the inference gate learn the hierarchical knowledge. The smaller HSC, the easier it is to activate the same experts for similar categories. 

\subsection{Adversarial Regularization}
Ideally different experts add different perspectives in the final ensemble score. In reality, however, experts tend to reach similar or identical conclusions in prediction tasks, especially if they see the same sequence of training data. To overcome this undesirable unanimity problem, 
we add a regularization term, formula~\eqref{eq:adv_loss}, that deliberately encourages disagreement among the experts. 

More specifically, for each input example in the MoE network, some experts are left idle, due to their relatively low gating values. Intuitively, the model determines that their predictions are somewhat irrelevant to the ground truth. 
Out of these, we randomly sample $D \leq N - K$ adversarial experts, whose indices are denoted by $U_D$, and subtract the L2 difference between their prediction probabilities and those of the top $K$ experts from the training loss. In other words, we reward those adversarial experts who predict differently from the top K experts.
As a mnemonic convention, these adversarial experts will also be called \textbf{disagreeing experts}. Note that $U_d(x^{sc}) \cap U_{top\_K} (x^{sc}) = \emptyset$.




We also observe that different examples have different sets of random disagreeing experts and top K inference experts, making the implementation less than straightforward. We define adversarial loss to measure the distance between the to joint expert sets for a single example as follows:
\begin{align}
    \label{eq:adv_loss}
    \advloss(X, x^{sc}) = \sum_{\substack{i \in U_{top\_K(x^{sc})}\\ j \in U_d(x^{sc})}} (\sigma(E_i(X)) - \sigma(E_j(X)))^2.
\end{align}



The $\advloss$ helps experts stay different from each other without directly interfering with the active expert predictions in the original MoE framework, the larger $\advloss$, the further the distance between disagreeing experts and inference experts.


\subsection{Combined Training Loss}

Our best model combines both Hierarchical Soft Constraint and Adversarial Loss during training. The full objective function contains three parts: (1) Cross Entropy Loss with respect to the user purchase binary label $y$, (2) HSC between the inference gates $G^I$ and constraint gates $G^C$, and (3) $\advloss$ between inference experts and disagreeing experts:

\begin{align}
    CE &= -(y \log \hat{y} + (1 - y) \log (1 - \hat{y})) \\
    \label{eq:loss_fn}
    J(\Theta) &= \frac{1}{n} \sum (CE
    + \lambda_1 HSC(x^{sc}, x^{tc})
    - \lambda_2 \advloss(X, x^{sc}))
\end{align}

Here  $n$ is the number of examples, $\lambda_1$ and $\lambda_2$ control the relative importance of corresponding item, for which we perform grid search in powers of 10. 

While the inference gating weights $\Theta_g$ affect all three components of the training loss, the expert tower weights $\Theta_{e_i}$, $1 \le i \le N$, do not affect the HSC regularization component. In other words, $\nabla_{\Theta_{e_i}} HSC \equiv 0$. Thus the gradient descent update formula simplifies slightly as follows:
\begin{align}
    \label{eq:o_e}
    \Theta_{e_i}^* &= \Theta_{e_i} - \nabla_{\Theta_{e_i}}(CE - \lambda_2 \advloss), \qquad 1 \le i \le N \\
    \label{eq:o_g}
    \Theta_g^* &= \Theta_g - \nabla_{\Theta_g}(CE + \lambda_1 HSC - \lambda_2 \advloss)
\end{align}

\section{Experiments}
In this section, we compare our improved MoE models with several MoE baselines empirically. We also report the improvements of various MoE-based models on test datasets with different categories in Section~\ref{section:performance-on-different-categories}. In Section~\ref{section:hyper-parameter-study}, we study the impact of different hyper-parameters on our model.
Towards the end (Section~\ref{section:onling_test}), we conduct the online A/B testing on the real traffic of active user on the JD e-commerce search system. 

\subsection{Experiment Setup}

To verify the effectiveness of our proposed model in a real industrial setting, we experiment on an in-house dataset. In addition, we run similar comparison experiments on a subset of the amazon review public dataset that covers all categories.


\subsubsection{In-House Dataset}
We collect users' purchase records from the JD e-commerce search system. Each example consists of product features (e.g., category, title, price), user features (e.g., age, behavior sequence), and query. In addition there are so-called 2-sided features that depend on both query/user and the product, for instance, the historical CTR of the product under the present query. 

The category system has a hierarchical tree structure, with the parent nodes given by the top-categories (TC) and child nodes by the sub-categories(SC). Data statistics are presented in Table~\ref{tab:data_statistics}.

\begin{table}
    \centering
    \caption{In-house dataset statistics.}
    \begin{tabular}{llrr}
        \hline
        \multicolumn{2}{c}{Statistics} &  Training Set & Test Set \\
        \hline
        \multirow{4}*{Data Size} & Complete & 26,674,871 & 2,059,293 \\
        ~ & Clothing(C) & 755,659 & 24,588 \\
        ~ & Books (B) & 1,520,243 & 75,218 \\
        ~ & Mobile Phone (M) & 1,344,726 & 73,549\\
        \hline
        \multirow{2}*{Category} & \#\ of Top Categories & 38 & 37 \\
        ~ & \#\ of Sub Categories  & 3,479 & 2,228\\
        \hline
        \multirow{2}*{Query} & \#\ of queries & 2,234,913 & 63,172 \\
        ~ & \#\ of query/item pairs & 9,978,755 & 1,479,115 \\
        \hline
    \end{tabular}
    \label{tab:data_statistics}
\end{table}

\subsubsection{Amazon Review Public Dataset \cite{ni2019justifying}}
We join (via distributed mapreduce) the publicly available 5-core, meta, and review data files into a single file, where each row represents a single user/item review event. The dataset includes 199,298,798 instances with 13,727,767 users, 6,926,608 goods and 979,421 categories.  Besides the features in the three raw files, we compute additional ones based on each user's review history, as well aggregate review on each item:

\begin{itemize}
    \item raw features: reviewerID, asin (item id), top\_category,
    
    sub\_category (leaf category), title, brand  
    \item user history features: asin\_hist,  brand\_hist
    \item item aggregate ratings: asin\_overall\_cnt,
    
    asin\_overall\_cnt\_\{1,2,3,4,5\}
    \item reviewer aggregate ratings: reviewer\_overall\_cnt,
    
    reviewer\_overall\_cnt\_\{1,2,3,4,5\}
\end{itemize}

For instance, asin\_hist arranges past item ids reviewed by the current reviewer in chronological order. asin\_overall\_cnt stands for the total number of reviews the current item has received, and reviewer\_overall\_cnt\_x stands for the number of reviews with rating x, given by the current reviewer. 
Our code \footnote{Experiment code on amazon dataset is available on GitHub: \url{https://github.com/advhscmoe/adv\_hsc\_moe}} is publicly available.

Top\_category and sub\_category function similarly to the in-house dataset experiment. One key difference from the latter dataset, however, is the lack of queries in the amazon dataset, making it more suitable for recommendation problems.

All features in the first two categories above are represented by 64 dimensional embeddings.

\subsubsection{Evaluation Metrics}
 We use two evaluation metrics in our experiments: AUC (Area Under the ROC Curve) and NDCG (Normalized Discounted Cumulative Gain) \cite{jarvelin2002cumulated}. AUC intuitively measures the agreement between the model and user purchase actions, on pairs of items within a session. 
 NDCG is a ranking metric that achieves similar effect, except it places more weight on pairs whose positions are close to the top of the search result page. On the in-house dataset, both metrics are computed on a per session basis and averaged over all sessions. Following \cite{liu2020kalman}, we use AUC for evaluation on the amazon review public dataset.

\subsubsection{Model Comparison}
 We compare 5 models in our experiments: DNN, MoE,
 Adversarial MoE (Adv-MoE), Hierarchical Soft Constraint MoE (HSC-MoE) and our model with both Adversarial experts and Hierarchical Soft Constraint (Adv \& HSC-MoE).

\subsubsection{Parameter Settings}
In our experiments, the DNN and a single expert tower have the same network structure as well as embedding dimension. 
For experiments on the in-house dataset, all sparse features are represented by 16 dimensiona embeddings and the single expert have 4 layers with dimension 1024, 512, 256, 1. For experiments on amazon public dataset, we follow the parameter settings(DNN network structure and batch size) in \cite{liu2020kalman}.
We use ReLU as the activation functions for hidden layer and AdamW proposed in \cite{loshchilov2017decoupled} as optimizer for all models. The learning rate is $10^{-4}$ for all models, $\lambda_1$ and $\lambda_2$ in objective function \eqref{eq:loss_fn} are both $10^{-3}$. To be fair, we use the same setting on MoE-based model (MoE, Adv-MoE, HSC-MoE, and Adv \& HSC-MoE), hyper-parameter including $num\_expert = 10$, $K = 4$. Adv-MoE and Adv \& HSC-MoE add a disagreeing expert $D = 1$ to calculate adversarial loss.

\subsection{Full Evaluations}
\subsubsection{Results on In-House Dataset}
The evaluation results for different models on in-house dataset are shown in Table~\ref{tab:performance-evaluation}. 

Compared with DNN model, our model(Adv \& HSC-MoE) achieves absolute AUC gain of 0.96\% (0.99\% in term of NDCG), which indicates its good generalization performance. We also have the following observations by carefully comparing the effects of different models.

\begin{itemize}
    \item All MoE-based networks improve upon DNN baseline model on all 4 metrics, including AUC, NDCG, and the both metrics restricted to the top 10 shown positions(AUC@10 and NDCG@10). The original MoE model already brings 0.44\% improvement on AUC and 0.55\% improvement on NDCG over DNN. 
    \item Hierarchical Soft Constraint has a stable improvement. It brings 0.19\% AUC (0.25\% NDCG) gain for HSC-MoE over MoE, and 0.34\% AUC (0.45\% NDCG) gain for Adv \& HSC-MoE over Adv-MoE.
    \item Using adversarial loss during model training also improves the model's generalization performance. This component brings an additional 0.18\% improvement in AUC compared to the HSC-MoE models.
\end{itemize}

\begin{table*}
    \centering
    \caption{Results on in-house dataset. Larger AUC and NDCG mean better performance. Numbers marked with $\ast$ are p values for DNN model. $\ddagger$ indicates the p value for MoE model.}
    \begin{tabular}{c||c|c|c|c||c|c|c|c}
        \hline
        ~ & \multicolumn{4}{c||}{metrics} & \multicolumn{4}{c}{p value} \\
        \hhline{-||----||----}
         Model & AUC@10 & AUC & NDCG@10 & NDCG & AUC@10 & AUC & NDCG@10 & NDCG \\
         \hhline{-||----||----}
         DNN & 0.6965 & 0.8131 & 0.5529 & 0.5820 & - & - & - & - \\
         \hhline{-||----||----}
         MoE & 0.7026 & 0.8175 & 0.5590 & 0.5875 & $\approx 10^{-5\ast}$ & $<10^{-20\ast}$ & $\approx 10^{-5\ast}$ & $\approx 10^{-5\ast}$ \\
         \hhline{-||----||----}
         Adv-MoE & 0.7027 & 0.8178 & 0.5592 & 0.5878 & $\approx 0.06{\ddagger}$ & $\approx 0.02^{\ddagger}$ & $\approx 0.03^{\ddagger}$ & $\approx 0.02^{\ddagger}$ \\
         \hhline{-||----||----}
         HSC-MoE & 0.7058 & 0.8194 & 0.5619 & 0.5900 & $\approx 0.01^{\ddagger}$ & $\approx 0.05^{\ddagger}$ & $\approx 0.02^{\ddagger}$ & $\approx 0.01^{\ddagger}$ \\
         \hhline{-||----||----}
         Adv \& HSC-MoE & \textbf{0.7084} & \textbf{0.8212} & \textbf{0.5645} & \textbf{0.5923} & $\approx 10^{-5\ddagger}$ & $<10^{-20\ddagger}$ & $\approx10^{-5\ddagger}$ & $<10^{-20\ddagger}$\\
         \hline
    \end{tabular}
    \label{tab:performance-evaluation}
\end{table*}

\subsubsection{Results on Amazon Review Public Dataset}
We also exam the contribution of our method to the amazon review public dataset. Table \ref{tab:performance-evaluation-amazon} shows the results of different models. We observe significant improvement with small p values from vanilla MoE model to advanced MoE models. The gains from our improved MoE models, Adv-MoE, HSC-MoE and Adv \& HSC-MoE, are consistent with in-house experiments, and validate the general applicability of adversarial regularization and hierarchical soft constraint techniques.

\begin{table}
    \centering
    \caption{Results on Amazon review public dataset \cite{ni2019justifying}. Numbers marked with $\ast$ are p values compared against DNN model. $\ddagger$ indicates the p value against MoE model.}
    \begin{tabular}{c|c|c}
        \hline
        Model & AUC & p value \\
        \hline
         DNN & 0.7031 & - \\
        \hline
         MoE & 0.7096 & $<10^{-20\ast}$ \\
        \hline
         Adv-MoE & 0.7314 & $<10^{-20\ddagger}$ \\
        \hline
         HSC-MoE & 0.7327 & $<10^{-20\ddagger}$ \\
        \hline
         Adv \& HSC-MoE & \textbf{0.7402} & $<10^{-20\ddagger}$ \\
        \hline
    \end{tabular}
    \label{tab:performance-evaluation-amazon}
\end{table}

\subsection{Performance on different categories}
\label{section:performance-on-different-categories}

 We test model performance on various categories to verify the benefit of our model in different categories of products. 
 Firstly, we evaluate the performance of our model in different categories, which have different data sizes in the in-house training dataset.
 Then, we use three categories datasets to train and compare DNN and our model. Finally, we analyze the distribution of inference MoE gate values in all categories to investigate the relationship between experts and categories.
 
\begin{figure}
    \centering
    \includegraphics[width=\linewidth]{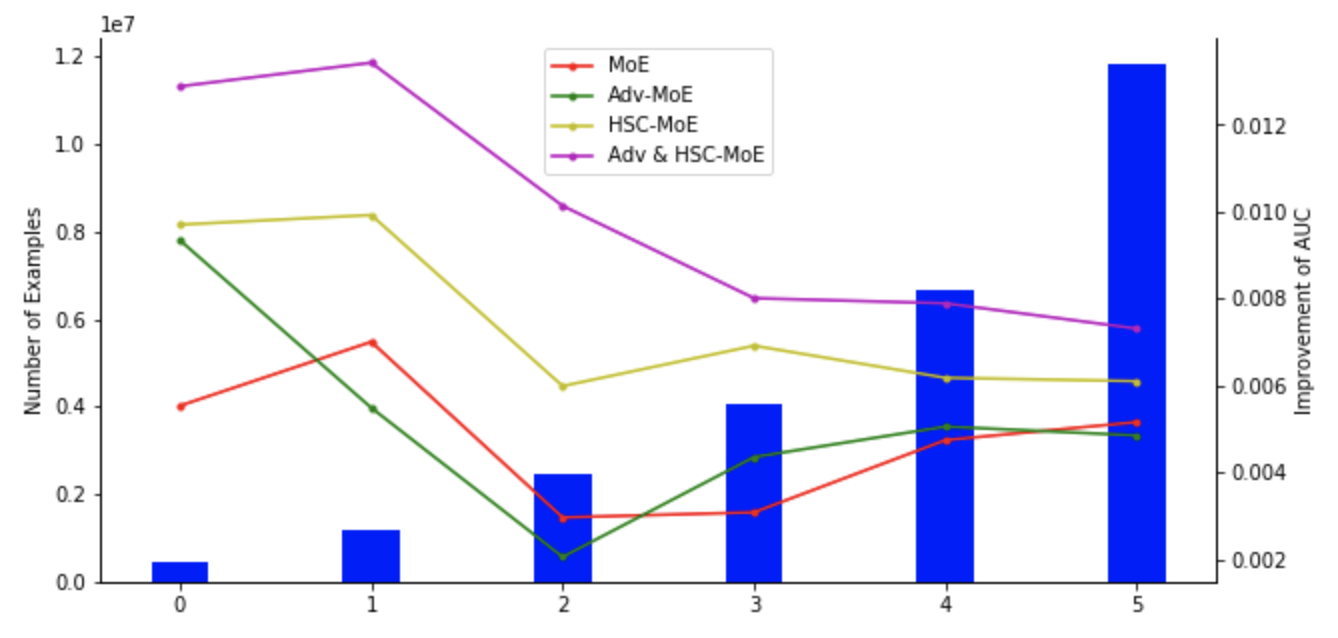}
    \caption{Model performance on different top categories. The X-axis corresponds to different category buckets. Left-hand Y-axis gives the combined data sizes of each category bucket, while the right-hand Y-axis shows AUC improvement with respect to the DNN baseline.}
    \label{fig:category-improvement}
\end{figure}

 We put various categories in different buckets according to data size in the training dataset. As presented in Figure~\ref{fig:category-improvement}, 
 the blue bar in the figure shows that the data size corresponding to categories bucket, left bar stands for some small categories with few training data, right is large categories otherwise. The left-hand side Y-axis corresponds to the data size of categories buckets.
 The right-hand side Y-axis is the improvement of AUC. Lines with different colors illustrate the improvement of AUC in different models with increasing data sizes in various categories bucket. All MoE-based models outperform the baseline model (DNN), as the AUC improvement are all greater 0 in all lines. 
 It is worth noting that our model (purple line) is more effective for small categories than for large categories, as the decreasing trend when from left to right. 
 
 It is likely that the improvement in small categories is owing to the HSC. The HSC constraint the distribution of gate value for different categories, which in turn affects the choice of experts. Similar categories are easier to activate the same experts and small categories can easier transfer learning from shared experts. 

 Moreover, we collect different training and testing datasets within three different categories from our e-commerce search log. The data statistics of the three categories are shown in Table~\ref{tab:data_statistics}, including Mobile Phone (M), Books (B), and Clothing (C). The dataset sizes of Books and Mobile Phone are sufficient to train a good model, thus the AUC for Books and Mobile Phone are always higher than that for Clothing. 
 As shown in Table~\ref{tab:3cid_model}, we train four versions of DNN models and an Adv \& HSC-MoE model as follows:
 
 \begin{itemize}
     \item \textbf{M-DNN} uses the training dataset of Mobile Phone category only to train a 3-layer Feed-Forward model.
     \item \textbf{B-DNN} uses the training dataset of Books category only.
     \item \textbf{C-DNN} uses the training dataset of Clothing category only.
     \item \textbf{Joint-DNN} use the joint dataset (A+B+C) to train DNN.
     \item \textbf{Joint-Ours} also uses the joint dataset (A+B+C) to train our best candidate model, namely HSC \& Adv-MoE.
 \end{itemize}

As shown in Table~\ref{tab:3cid_model}, we test those models on all 3 category test sets separately.
 The joint training dataset is more beneficial for those categories with less training dataset. 
 It can be seen that there is 0.36\% improvement of AUC in Clothing, compared with 0.25\%, and -0.08\% AUC gain in Books and Mobile Phone. 
 Meanwhile, our method (Joint-Ours) outperforms Joint-DNN and separate-DNNs in all categories, showing the advantage of our proposed method in different categories. The improvement of AUC are the same as Figure~\ref{fig:category-improvement}, where smaller categories (e.g. Clothing) gain a higher improvement. 

\begin{table}
    \centering
    \caption{Evaluations on different training and testing datasets. M, B, C are datasets on three categories respectively. M: Mobile Phone, B: Books, C: Clothing}
    \label{tab:3cid_model}
    \begin{tabular}{c|c|c|c|c}
    \hline
        \multirow{2}*{Model} & \multirow{2}*{Train set} & \multicolumn{3}{c}{Test set(AUC)} \\
        \cline{3-5}
        ~ & ~ & M & B & C \\
        \hline
        M-DNN & M & 0.8059 & - & - \\
        \hline
        B-DNN & B & - & 0.8393 & - \\
        \hline
        C-DNN & C & - & - & 0.7957 \\
        \hline
        Joint-DNN & \multirow{2}*{M + B + C} & 0.8051 & 0.8418 & 0.7993 \\
        \cline{1-1}
        \cline{3-5}
        Joint-Ours & ~ & \textbf{0.8098} & \textbf{0.8422} & \textbf{0.8052} \\
        \hline
    \end{tabular}
\end{table}

 In order to clearly investigate the impact of HSC and adversarial loss, we analyze the distribution of inference MoE gate values in all categories. In our experiments, inference MoE gate values form a $10$-dimensional vector, which stands for the probability that each expert should be activated, for a given example. To clearly illustrate the relationship between categories and activated experts, we cluster those gate values into 2-dimension using t-SNE \cite{maaten2008visualizing}, which effectively learns 2-dimensional points that approximate the pairwise similarities between output vectors of the gate network for a set of input examples. We group together 
semantically similar categories and assign a distinct color to each group as shown in the Table~\ref{tab:cid_class}.

\begin{figure}
    \centering
    \subfigure[MoE]{
        \begin{minipage}[t]{0.5\linewidth}
            \centering
            \includegraphics[width=\linewidth]{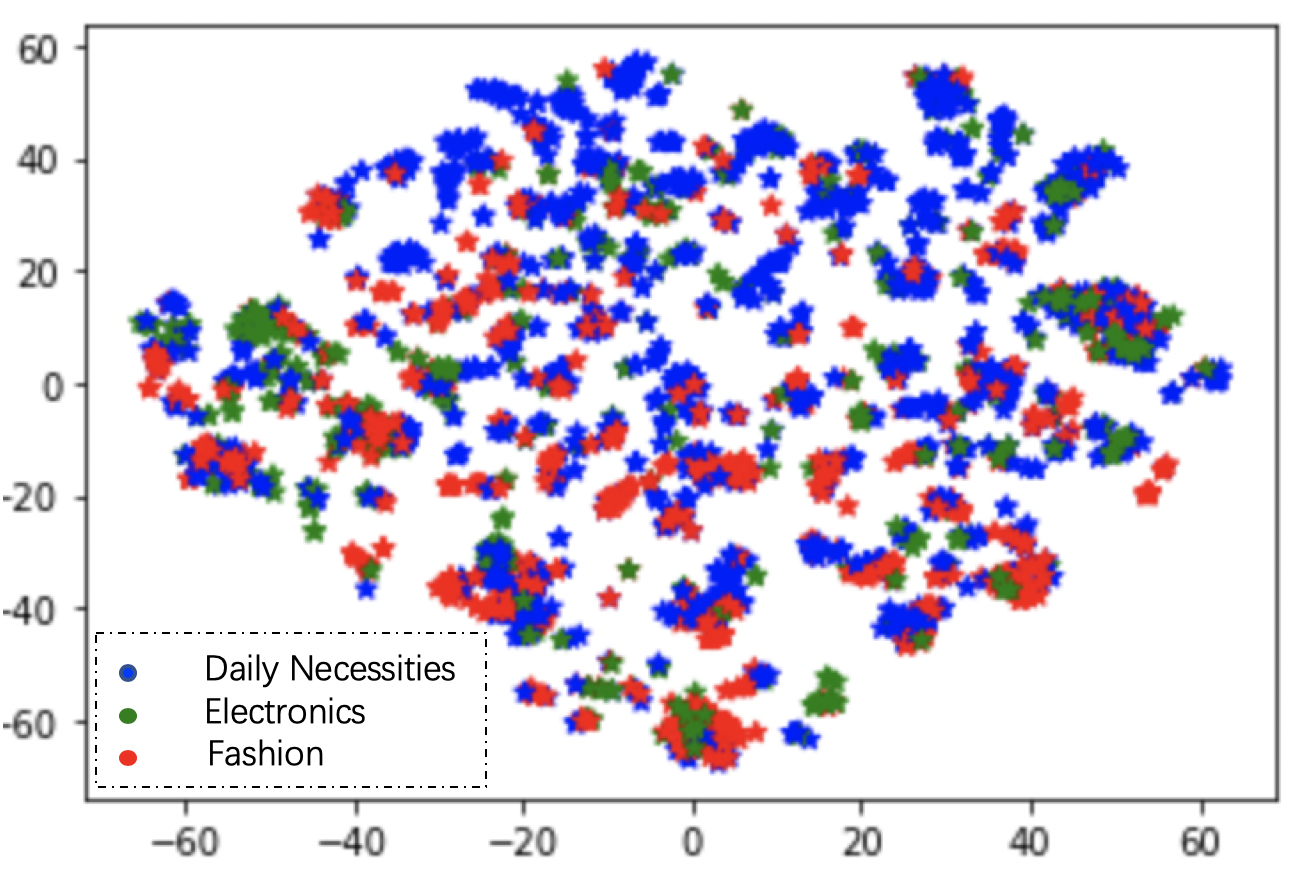}
            \label{fig:moe-gatevalue}
        \end{minipage}%
    }%
    \subfigure[Adv-MoE]{
        \begin{minipage}[t]{0.5\linewidth}
            \centering
            \includegraphics[width=\linewidth]{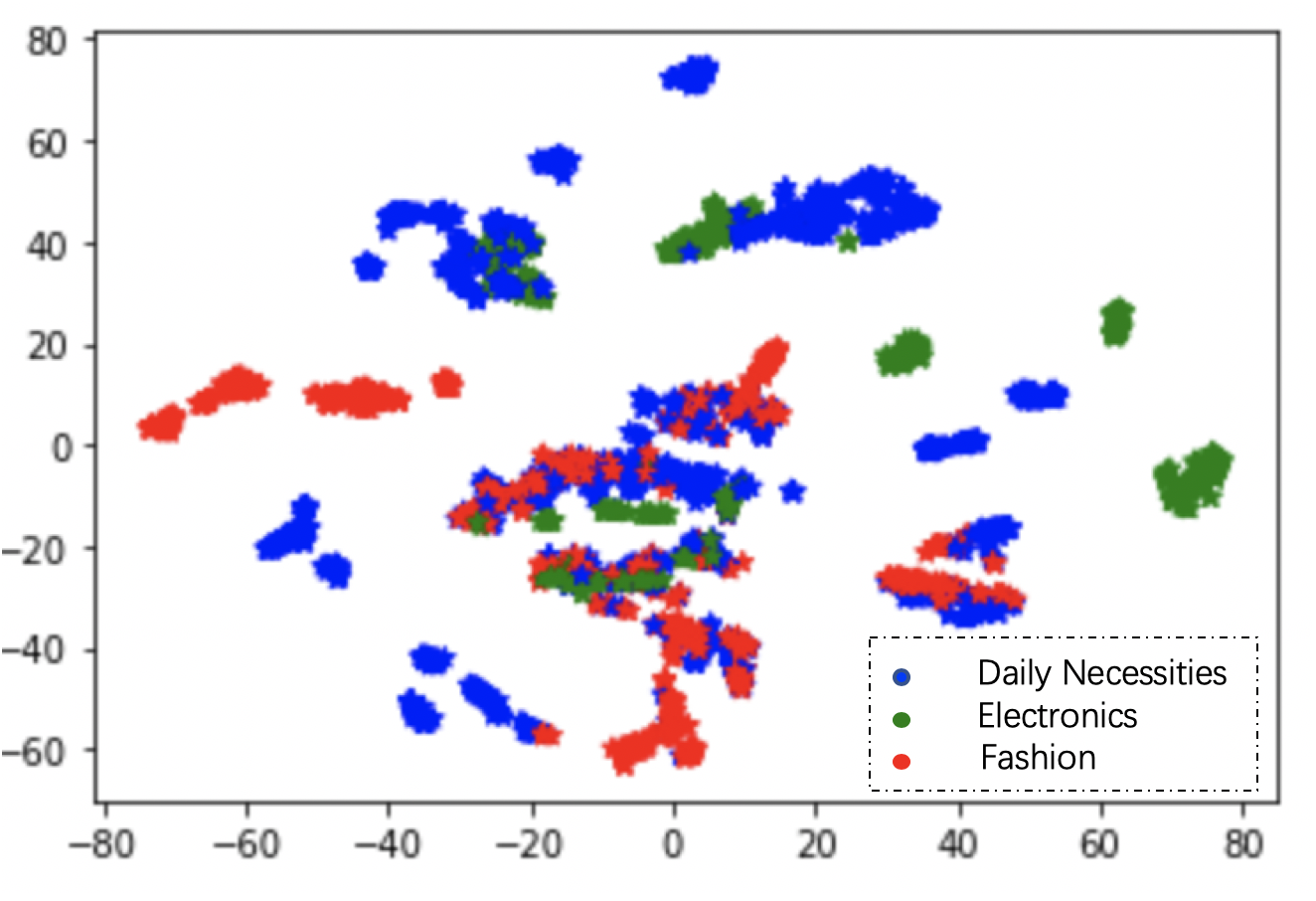}
            \label{fig:adv-gatevalue}
        \end{minipage}%
    }%
    
    \subfigure[Adv \& HSC-MoE]{
        \begin{minipage}[t]{0.5\linewidth}
            \centering
            \includegraphics[width=\linewidth]{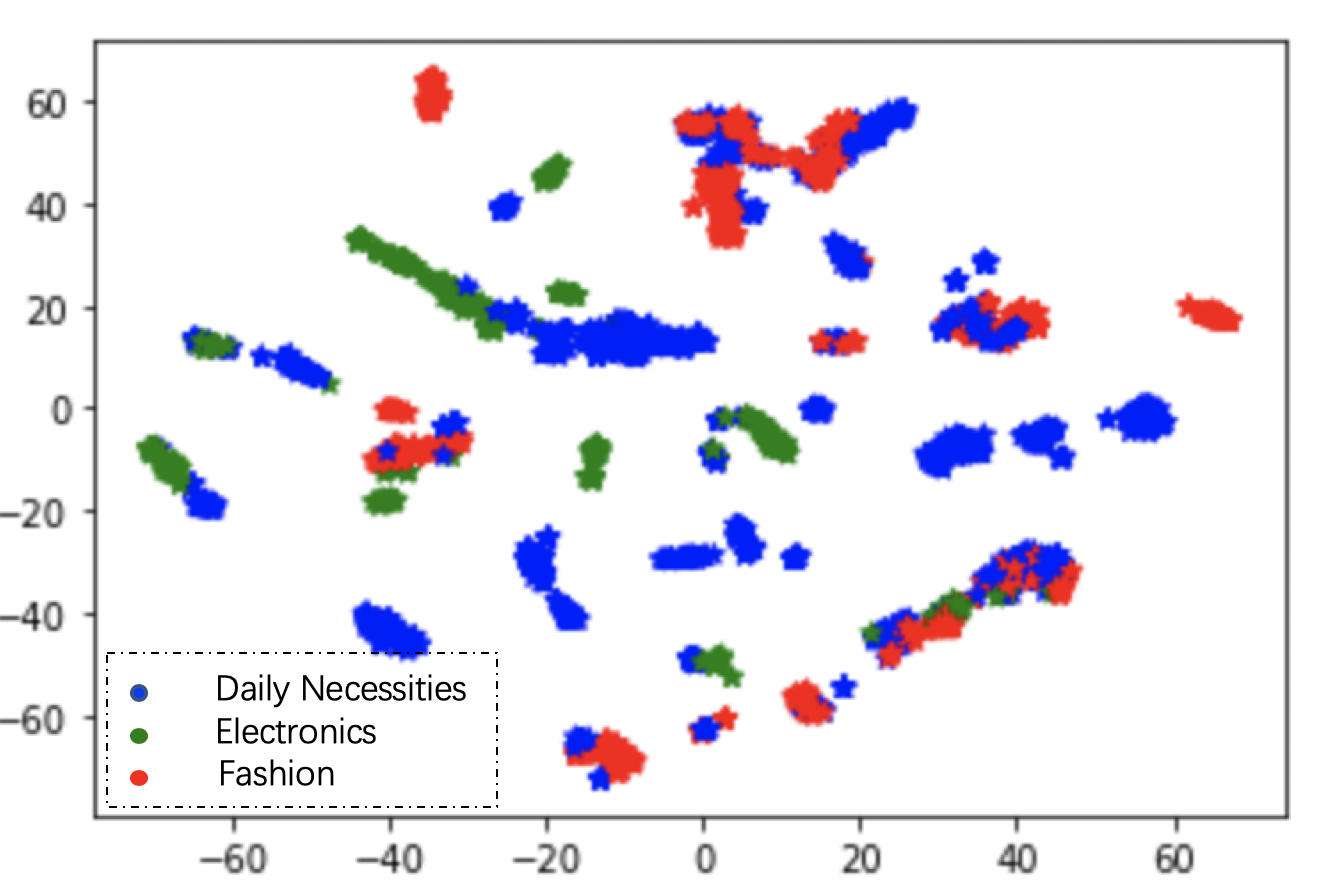}
            \label{fig:adv-hsc-gatevalue}
        \end{minipage}%
    }%
    \centering
    \caption{Distribution of inference gate values in different models. Points with the same color appear better clustered under our improved MoE models, indicating that similar categories are better able to share similar sets of experts.}
    \label{fig:gate_value}
\end{figure}

 Figure~\ref{fig:gate_value} makes it clear that similar categories have much more similar gate vectors under Adv-MoE and Adv \& HSC-MoE than under the vanilla MoE. In particular, semantically similar categories form much more structured clumps in
 Figure~\ref{fig:adv-hsc-gatevalue} and Figure~\ref{fig:adv-gatevalue} than the MoE baseline in Figure~\ref{fig:moe-gatevalue}.

 Moreover, between Figure~\ref{fig:adv-hsc-gatevalue} and Figure~\ref{fig:adv-gatevalue}, the presence of HSC gates produces an even cleaner separation of clusters than adversarial regularization alone. 
 
\begin{table}
    \centering
    \caption{Coloring scheme of similar category grouping}
    \begin{tabular}{c|c|c}
         \hline
        Semantic Class & Color & Representative Categories \\
        \hline
        Daily Necessities & blue & Foods, Kitchenware, Furniture $\dots$ \\
        \hline
        Electronics & green & Mobile Phone, Computer $\dots$ \\
        \hline
        Fashion & red & Clothing, Jewelry, Leather $\dots$ \\
        \hline
    \end{tabular}
    \label{tab:cid_class}
\end{table}

\subsection{Hyper-Parameter Study}
\label{section:hyper-parameter-study}

 We test different hyper-parameter settings in our model, including 1) the number of experts $N$ and different number of top experts $K$ and number of disagreeing experts $D$; 2) different input features for the MoE gate network; 3) the weight multipliers $\lambda_1$ and $\lambda_2$ for HSC and AdvLoss in the training objective function.

\begin{figure}
    \centering
    \includegraphics[width=\linewidth]{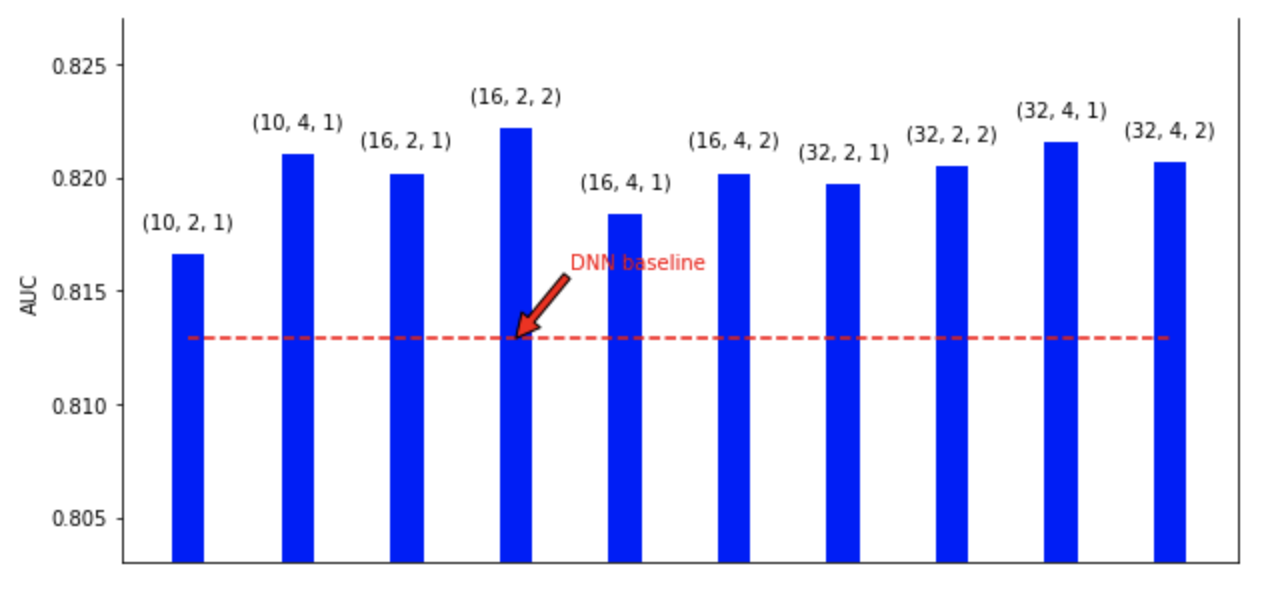}
    \caption{The HSC \& Adv-MoE model under different ($N$, $K$, $D$) hyper-parameter settings. }
    \label{fig:different_auc}
\end{figure}

 We set the total number of experts $N$ to be 10, 16, 32; the number of chosen experts $K$ to be 2, 4; and the number of adversarial experts $D$ to be 1, 2. As presented in Figure~\ref{fig:different_auc}, holding the other parameters fixed, increasing $K$ consistently improves model generalization. This is expected since higher $K$ yields greater expert capacity per example. On the other hand, there is no monotonic pattern among the other parameters $N$ and $D$, as evidenced by the pairs of triplets (16, 2, 2), (32, 2, 2), and (32, 4, 1), (32, 4, 2) respectively. A very large $N$ would dilute the amount of training data seen by each expert. Too many adversarial experts ($D \gg 0$) can also prevent the overall model from learning even the basic notions of relevance.
 Overall, our combined candidate model (HSC \& Adv-MoE) achieves the best test AUC on our search log dataset when $N = 16$, $K = 2$ and $D = 2$. 

 We test different gate input features in our model. As presented in Table~\ref{tab:gate_feature}, the model is able to achieve the best performance when using sub-categories alone. Adding top-categories, query, user feature or even all features does not bring benefit. 
 
 One possible explanation is that adding other features brings some noise that activates the ``wrong" experts for special categories. 
 The model gets the worst performance when we feed all features to the inference gate. The use of all features, including some product specific features, causes different products in the same query session to have different inference gate values. This causes variance between expert towers to dominate the variance between intra-session products, leading to ranking noise. The result demonstrates that the inference gate should be fed query-side features only in our model to guarantee a unique experts set as well as weight values within the same query session.

\begin{table}
    \centering
    \caption{The model performance in different gate input feature; the other hyper-parameters remain the same: $N = 10$, $K = 4$, $D = 1$, $\lambda_1 = \lambda_2 = 0.01$.}
    \label{tab:gate_feature}
    \begin{tabular}{c|c}
        \hline
        gate input feature &  AUC \\
        \hline
        SC & \textbf{0.8212} \\
        \hline
        (TC, SC) &  0.8137 \\
        \hline
        (query, TC, SC) &  0.8135\\
        \hline
        (user feature, TC, SC) & 0.8131 \\
        \hline
        all features & 0.8129 \\
        \hline
    \end{tabular}
\end{table}

\begin{table}[!htbp]
    \centering
    \caption{The test AUC in experiments with different combinations of $\lambda_1$ and $\lambda_2$. The other parameters remain the same: $N = 10$, $K = 4$, $D = 1$.}
    \label{tab:lambda}
    \begin{tabular}{c|c|c|c|c}
        \hline
        ~ & $\lambda_1$ = 0.1 & $\lambda_1$ = 0.01 & $\lambda_1$ = $10^{-3}$ & $\lambda_1$ = $10^{-4}$\\
        \hline
        $\lambda_2$ = 0.1 & 0.8167 & 0.8140 & 0.8167 & 0.8177 \\
        \hline
        $\lambda_2$ = 0.01 & 0.8217 & 0.8212 & 0.8172 & 0.8214 \\
        \hline
        $\lambda_2$ = $10^{-3}$ & 0.8221 & 0.8168 & \textbf{0.8227} & 0.8221 \\
        \hline
        $\lambda_2$ = $10^{-4}$ & 0.8217 & 0.8223 & 0.8225 & 0.8172 \\
        \hline
    \end{tabular}
\end{table}
\subsection{Experiments On Online A/B Testing}
\label{section:onling_test}

The parameters $\lambda_1$ and $\lambda_2$ in the objective function \eqref{eq:loss_fn} control the relative importance of $HSC$ and $\advloss$ in the overall training loss. 
We make parameter sweeps from $10^{-4}$ to $0.1$ for both $\lambda_1$ and $\lambda_2$. As shown in Table~\ref{tab:lambda}, our model achieves the best performance when $\lambda_1 = 10^{-3}$ and $\lambda_2 = 10^{-3}$.


We conducted online A/B testing on the JD e-commerce search engine over a six day period.
Compared to the DNN model, Our model has improved CVR by 2.18\% and RPM(Revenue Per Mille) by 3.97\%.

\section{Conclusions and Future Work}
The adversarial regularization and hierarchical soft constraint techniques presented here are promising steps towards developing a category-aware ranking model in product search. It achieves significant improvement on an industry scale dataset, mainly from these advantages: 1) Small sub-categories under the same top-category are able to share similar experts, thereby overcoming parameter sparsity under limited training data. 2) Adversarial regularization encourages the experts to ``think independently" and approach each problem from a diversity of angles. 

A few directions remain to be explored. First, we did not use product side category id as input to the gate network, since the result actually deteriorates compared to using only query-side category information. One explanation is that adding item side gating input causes different items to go through different experts under the same query, leading to large prediction variance. Product side categories are typically more accurate, however, and we plan to incorporate them by exploring more factorized architectures, with multi-phased MoE.

The 2-level category hierarchy considered here can be viewed as the simplest form of knowledge graph. An interesting generalization is to apply the soft constraint technique to more general human or model annotated knowledge graph, provided the latter has enough coverage on the training corpus.

Another important source of features in e-commerce search deals with personalization, in particular user historical behavior sequence. While our techniques do not require such structured data, it is natural to apply different experts to different events in the user history, and hopefully focus more on the relevant historical events.

Lastly it is desirable to fine-tune individual expert models to suit evolving business requirement or training data. Thus it would be interesting to assess transfer learning potential based on the component expert models.

\section{appendix}
\subsection{Category Inhomogeneity}

We discuss variance of features across categories in our training data, as a motivation for dedicated expert tower combinations for different categories.
 
Let $I$ be the ranked product list in a search session, and let $i^a, i^b, \ldots \in I$ denote items in $I$. Given a feature $f$, we define its \textbf{feature-importance} to be the ROC-AUC of the item ranking based on $f$, with respect to the user purchase label:
 \begin{equation}
    \label{eq:pif}
     FI(f) = \frac{1}{N}\sum \frac{\#\{(i^a, i^b): f_{i^a} > f_{i^b}\}}{\#\{(i^a, i^b) : y_{i^a} = 1, y_{i^b} = 0\}}.
 \end{equation}
Here $y_{i^a} = 1$ means the item $i^a$ has been purchased, $y_{i^b} = 0$ means the item $i^b$ has not been purchased, and $(i^a, i^b)$ range over item pairs in a single query session. The expression $\#(S)$ stands for the number of items in the set $S$. $N$ is the number of search sessions. 
 
\begin{figure}
    \centering
    \subfigure[Inter-categories]{
    \begin{minipage}[t]{0.5\linewidth}
        \label{fig:fi_inter}
        \centering
        \includegraphics[width=\linewidth]{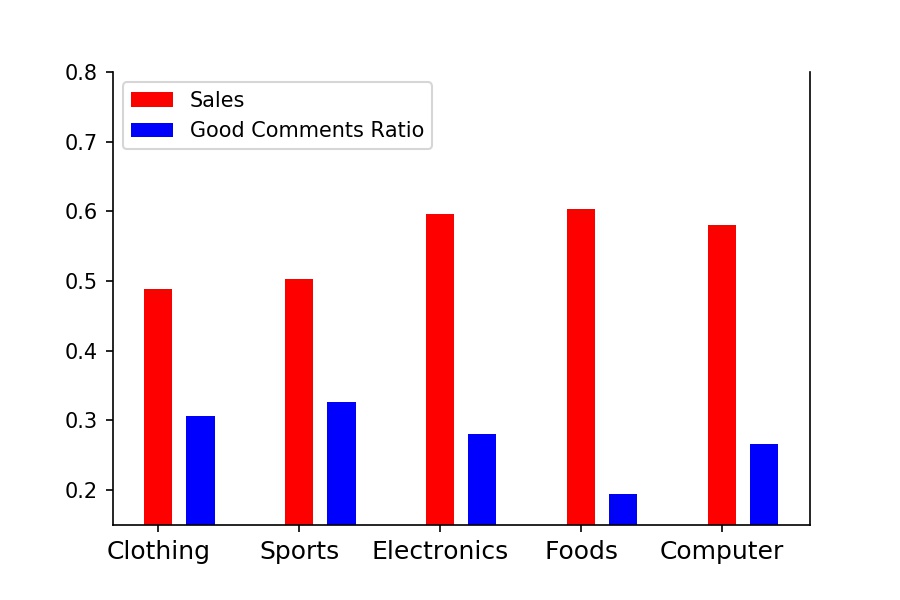}
    \end{minipage}%
    }%
    \subfigure[Intra-categories]{
    \begin{minipage}[t]{0.5\linewidth}
        \label{fig:fi_intra}
        \centering
        \includegraphics[width=\linewidth]{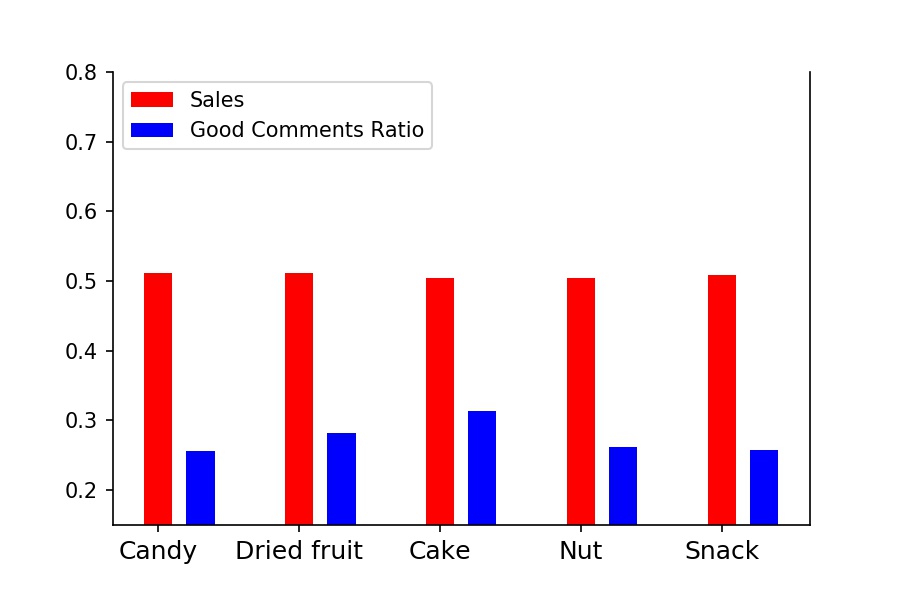}

    \end{minipage}%
    }%
    \caption{Feature-importance in different categories. The X-axis corresponds to different categories, and Different colors refer to different features.}
    \label{fig:fi}
 \end{figure}

 We analyze products in five different categories from our search logs: Clothing, Sports, Foods, Computer, and Electronics. 
 Figure~\ref{fig:fi_inter} shows the feature-importance of different features, including sales volume and good comments ratio, in different top-categories. 
 Good comments ratio is likely more important in Clothing and Sports products than in Foods, because users pay more attention to bad comments to avoid defective products in the Clothing or Sports categories. In Foods, Computer, or Electronics, they may tend to buy more popular products with high sales volume. We also compute the feature-importance in sub-categories within the same top-category, namely Foods(Figure~\ref{fig:fi_intra}). In contrast to the high variance of feature importance among top-categories, the intra-category feature-importances are more similar. Other top-categories have similar intra-category variance. This agrees with the intuition that users focus their attention to similar features when buying products from the same top-category. 
 
\begin{figure}
    \centering
    \subfigure[Inter-categories]{
    \begin{minipage}[t]{0.5\linewidth}
        \label{fig:inter_sale}
        \centering
        \includegraphics[width=\linewidth]{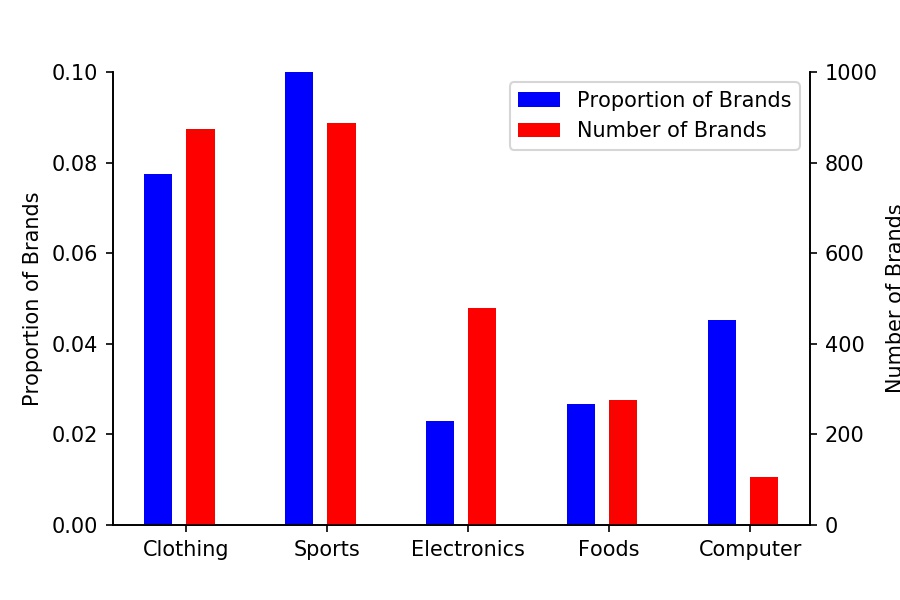}
    \end{minipage}%
    }%
    \subfigure[Intra-categories]{
    \begin{minipage}[t]{0.5\linewidth}
        \label{fig:intra_sale}
        \centering
        \includegraphics[width=\linewidth]{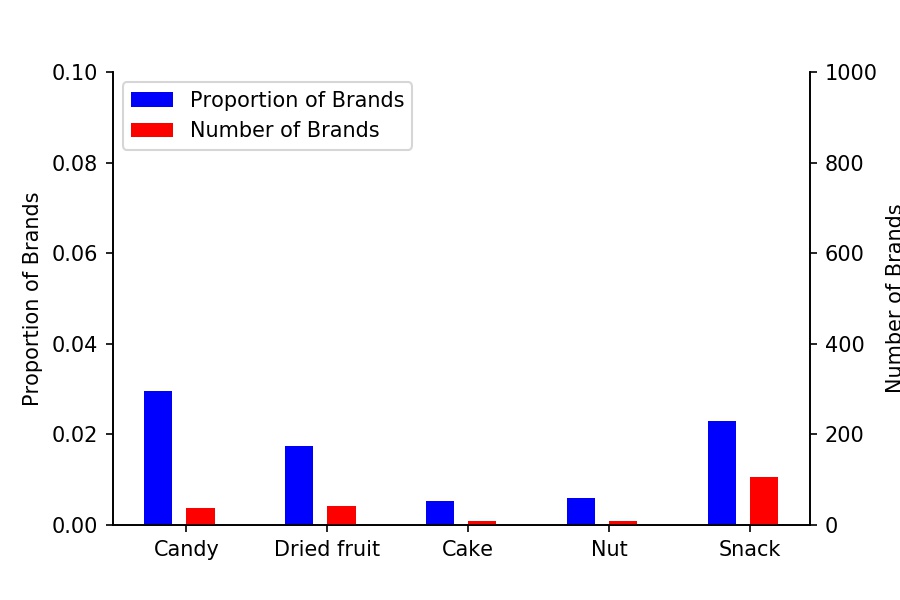}

    \end{minipage}%
    }%
    \caption{The Proportion and number of brands in the top 80\% of products by sale volume ranking in different categories. The X-axis refers to categories, the left-hand Y-axis corresponds to the Proportion of Brands, and the right-hand Y-axis corresponds to the Number of Brands.}
    \label{fig:sale_ratio}
 \end{figure}

 In order to assess the variance of sparse features, we examine the relationship between sparse features and the sales volume in specific categories. 
 We look at the distribution of the proportion and absolute number of brands (a sparse feature in ranking model) in top 80\% of products by sale volume ranking. As shown in Figure~\ref{fig:sale_ratio}, the X-axis corresponds to the categories, the left-hand Y-axis refers to the proportion of brands and the right-hand Y-axis refer to the number of brands. The sales volume in Electronics are concentrated in the top brands, as top 80\% of sales in top 2\% brands. It means that the top brands have a great influence when users decide whether to buy a product in Electronics. In contrast, 
the distribution of Sports brand is more dispersed than Electronics, as top 80\% of product sales are scattered in nearly 10\% brands. Similar to the feature-importance, we also compute the top 80\% of sales in sub-categories within top-categories of Foods. The resulting intra-category variance is significantly smaller than inter-categories as shown in Figure~\ref{fig:intra_sale}. Consistent with the observation regarding numeric features feature-importance, sparse features have wildly differing influences on the purchase decision of inter-categories products, however, they have similar importance among sibling sub-categories.
  
  The observations we found in the log data verify the intuition that features, whether it is numeric or sparse, have different importance on the purchase decision of different categories. This motivates us to develop a category-wise model to capture different ranking strategies on different categories in product search.

\bibliography{sample-base}

\end{document}